\documentclass{article}

\usepackage{microtype}
\usepackage{graphicx}
\usepackage{subcaption}
\usepackage{multirow}
\usepackage{booktabs} % for professional tables

\usepackage{hyperref}

\usepackage[accepted]{icml2020}
\usepackage{authblk}

\icmltitlerunning{Benchmarking Differentially Private Residual Networks for Medical Imagery}

\begin{document}

\twocolumn[
\icmltitle{Benchmarking Differentially Private Residual  \\
           Networks for Medical Imagery}



\icmlsetsymbol{equal}{*}

\begin{icmlauthorlist}
\icmlauthor{Sahib Singh}{equal,zoo,noo,moo}
\icmlauthor{Harshvardhan D. Sikka}{equal,zoo,moo,goo}
\icmlauthor{Sasikanth Kotti}{zoo,foo}
\icmlauthor{Andrew Trask}{zoo,doo}
\end{icmlauthorlist}

\icmlaffiliation{zoo}{OpenMined}
\icmlaffiliation{moo}{Manifold Computing Group}
\icmlaffiliation{noo}{Ford R\&A}
\icmlaffiliation{goo}{Harvard University}
\icmlaffiliation{foo}{Indian Institute of Technology Jodhpur}
\icmlaffiliation{doo}{University of Oxford}

\icmlcorrespondingauthor{Sahib Singh}{sahibsin@alumni.cmu.edu}
\icmlcorrespondingauthor{Harshvardhan D. Sikka}{has727@g.harvard.edu}
\icmlcorrespondingauthor{Andrew Trask}{andrew@openmined.org}
\icmlkeywords{Machine Learning, ICML}

\vskip 0.3in]



\printAffiliationsAndNotice{\icmlEqualContribution} % otherwise use the standard text.

\begin{abstract}
In this paper we measure the effectiveness of $\epsilon$-Differential Privacy (DP) when applied to medical imaging. We compare two robust differential privacy mechanisms: Local-DP and DP-SGD and benchmark their performance when analyzing medical imagery records. We analyze the trade-off between the model's accuracy and the level of privacy it guarantees, and also take a closer look to evaluate how useful these theoretical privacy guarantees actually prove to be in the real world medical setting.

\end{abstract}
\section*{Introduction}
Hospitals and other medical institutions often have vast amounts of medical data which can provide significant value when utilized to advance research. However, this data is often sensitive in nature and as such is not readily available for use in a research setting, often due to privacy concerns. We seek to find privacy preserving mechanisms so that medical facilities can better utilize their data while also maintaining the degree of privacy required.

To facilitate secure model training we seek to compare two different approaches of Differential Privacy relevant to image data: Local-DP and DP-SGD.
We benchmark their performances over the following publicly available medical datasets:

1) \textbf{Chest X-Ray Images Pneumonia Detection Dataset} \cite{kermany2018identifying}:  The Chest X-Ray dataset consists of approximately 5,800 images sourced from chest radiography, which is used by medical specialists
to confirm pneumonia and other medical concerns though
they are not often the sole point of diagnosis. Different
radiographic images taken at separate time intervals, such as
before and during an illness, are often useful to physicians
during the diagnosis. In general, these images form up an
important part of an often multi-stage diagnosis process.
The percentage of deaths attributed to pneumonia and influenza
is 8.2\% in the United States, exceeding the threshold
of epidemic classification, which is 7.2\% \cite{cdc20}. Recently,
deaths due to pneumonia have sharply increased
due to the worldwide presence of COVID-19 and the
SARS-CoV-2 virus. The rapid construction and evaluation
of relevant models to track, diagnose, or support the
treatment and mitigation of the COVID-19 is critical given
global circumstances.

%%%%%%%
\begin{figure*}
    \centering
    \begin{subfigure}{0.15\textwidth}
     \includegraphics[width=\linewidth]{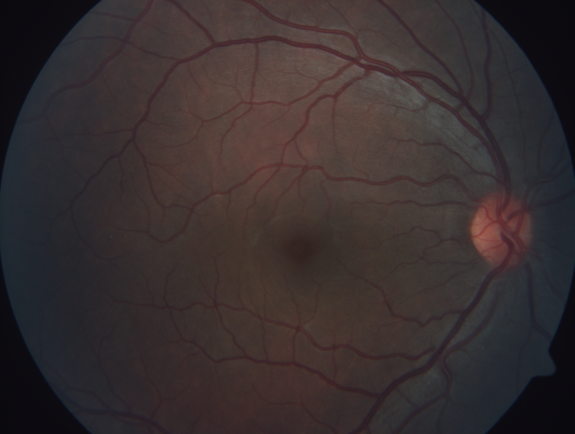}
     \caption{No DR}
     \label{fig:acc:high}
    \end{subfigure}
    \begin{subfigure}{0.15\textwidth}
     \includegraphics[width=\linewidth]{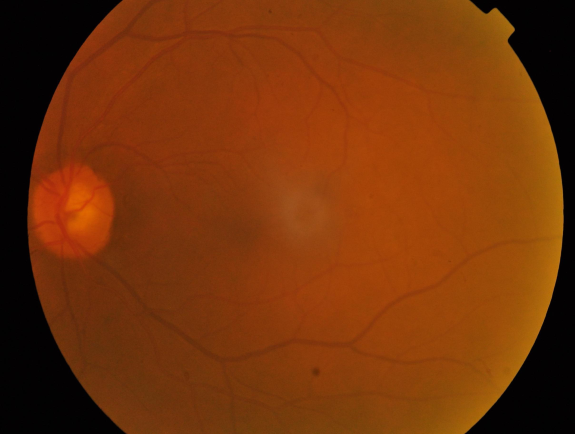}
     \caption{Mild DR}
     \label{fig:acc:non}
    \end{subfigure}
    \begin{subfigure}{0.15\textwidth}
     \includegraphics[width=\linewidth]{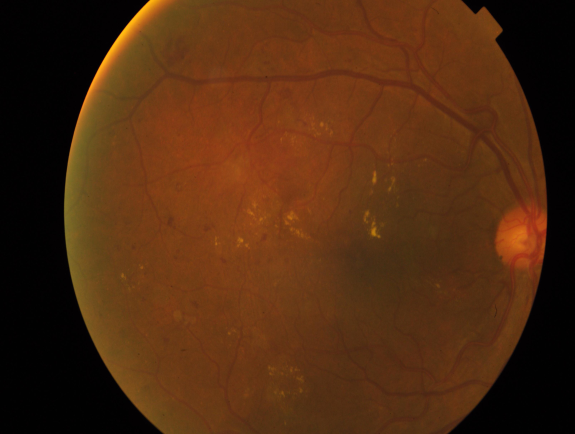}
     \caption{Moderate DR}
     \label{fig:acc:low}
    \end{subfigure}
    \begin{subfigure}{0.15\textwidth}
     \includegraphics[width=\linewidth]{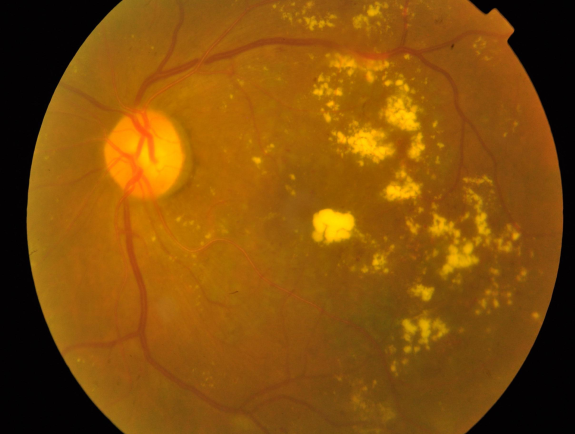}
     \caption{Severe DR}
     \label{fig:acc:high}
    \end{subfigure}
    \begin{subfigure}{0.15\textwidth}
     \includegraphics[width=\linewidth]{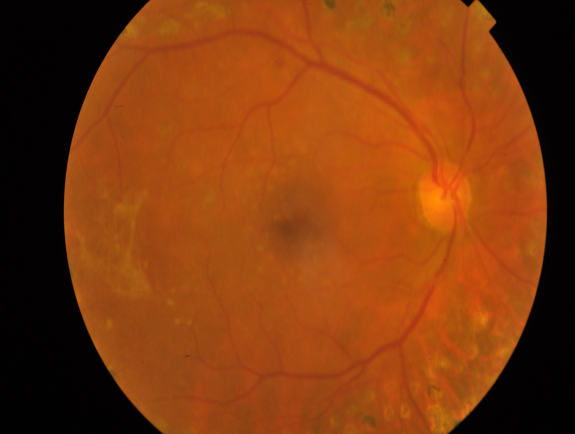}
     \caption{Proliferative DR}
     \label{fig:acc:high}
    \end{subfigure}
    \vspace{-1ex}
    \caption{Examples from the APTOS Blindness Detection Dataset. Samples progress in severity of Diabetic Retinopathy.}
    \label{fig:acc}
    \vspace{-1ex}
\end{figure*}

%%%%%%%%%%%

2) \textbf{APTOS 2019 Blindness Detection Dataset} \cite{atpos2019}: The dataset comprises of approximately 3600 training images of retina taken using fundus photography under a variety of imaging conditions. These retinal images can help identify diabetic retinopathy automatically.
Diabetic retinopathy (DR) refers to a diabetes complication affecting eyes. It's caused by damage to the blood vessels of the light-sensitive tissue at the back of the eye (retina). A clinician has rated each image for the severity of diabetic retinopathy on a scale of 0 to 4 where 0 is the case of No DR and 4 is Proliferative DR (Figure \ref{fig:acc}).  

Given the urgent need for these developments and the inherently sensitive nature of medical data, training and evaluating models while maintaining obfuscation
of critical personal information in the data corpus is vital.
This analysis of these DP mechanisms seeks to aid medical professionals better understand the tradeoff between accuracy and data privacy, and serve as a useful reference to better evaluate how sensitive information must be preserved while still ensuring the data remains useful for research purposes. Furthermore the analysis will also be useful to evaluate which kind of DP Mechanism would be more relevant for the use case. 

We start off by elucidating the related research previously conducted in the DP space, followed by an introduction to the DP mechanisms as they pertain to the our findings. We then evaluate the relevant experiments conducted. Finally, we discuss the significance of these findings and explore future directions.

\section*{Related Work}
It's worth noting that there are several techniques to achieve Differential Privacy for images. The DP techniques used in this experiment include Local-DP (LDP) which adds perturbations directly to the image, and DP-Stochastic Gradient Descent (DP-SGD) which achieves the privacy guarantees by adding noise to the gradient during the model training process.

Lately, there has also been research in applying Differential Privacy during the test-time inference stage in deep learning. These include mechanisms such as Cloak \cite{mireshghallah2020principled} and ARDEN \cite{wang2018not}. There are also well understood cryptographic protocols such as Homomorphic Encryption (HE) and Secure Multi-Party Computation (SMC) which could be used to encrypt data during training and inference phases. However the computational costs associated with these cryptographic techniques could be prohibitively expensive.
A comprehensive list of such privacy preserving techniques can be found in \citet{mirshghallah2020privacy}. 

We would additionally like to acknowledge similar work which was done earlier in a more theoretical setting \cite{fan2019differential}. Our paper builds upon the work and applies it towards the Healthcare domain, and is relevant in Medical Research in particular.
Other works relevant to general privacy preserving techniques for health records include \cite{kim2018privacy, kaissis2020secure, adam2007privacy, qayyum2020secure, dankar2013practicing, 8861857, faravelon2010towards}. Research related towards implementation techniques for secure deep learning include \cite{ryffel1811generic, dahl2018private, mireshghallahinterpretable}.

\section*{Preliminaries}
In this section we discuss the fundamental privacy preserving concepts used throughout the paper. 

\subsection*{Differential Privacy (DP).  \cite{dwork2006our,dwork2006calibrating}}

The central idea in differential privacy is the introduction of randomized noise to ensure privacy through plausible deniability. Based on this idea,
for  $\epsilon  \geq 0$, an algorithm $A$ is understood to satisfy Differential Privacy if and only if for any pair of datasets that differ in only one element, the following statement holds true.
\begin{center}
$P[A(D) = t] \leq e^\epsilon$ $P[A(D^{'}) = t] $ $\; \forall t$
\end{center}

Where $D$ and $D'$ are differing datasets by at most one element, and $P[A(D) = t]$ denotes the probability that $t$ is the output by $A$. This setting approximates the effect of individual opt-outs by minimizing the inclusion effect of an individual’s data. 

\subsection*{Local Differential Privacy (LDP). \cite{kasiviswanathan2011can}}
One major limitation of Differential Privacy is that the data owners will have to trust a central authority, i.e. the database maintainer, to ensure their privacy. Hence in order to ensure stronger privacy guarantees we utilize the concept of Local Differential Privacy \cite{bebensee2019local}.  
We say that an algorithm $\pi$ satisfies $\epsilon$-Local Differential Privacy
where $\epsilon > 0$ if and only if for any input $v \; and \;v^{'}$.
\begin{center}
$ \forall y \; \in Range(\pi) : P[\pi(v) = y] \leq e^\epsilon$ $P[\pi(v^{'}) = y]$
\end{center}

For $\epsilon-LDP$ the privacy loss is captured by $\epsilon$. Having $\epsilon = 0$ ensures perfect privacy as $e^{(0)} $ = 1, on the other hand, $\epsilon = \infty$ provides no privacy guarantee.
The choice of $\epsilon$ is quite important as the increase in privacy risks is proportional to $e^{\epsilon} $.

%%%%%%%
\begin{figure*}
    \centering
    \begin{subfigure}{0.2\textwidth}
     \includegraphics[width=\linewidth]{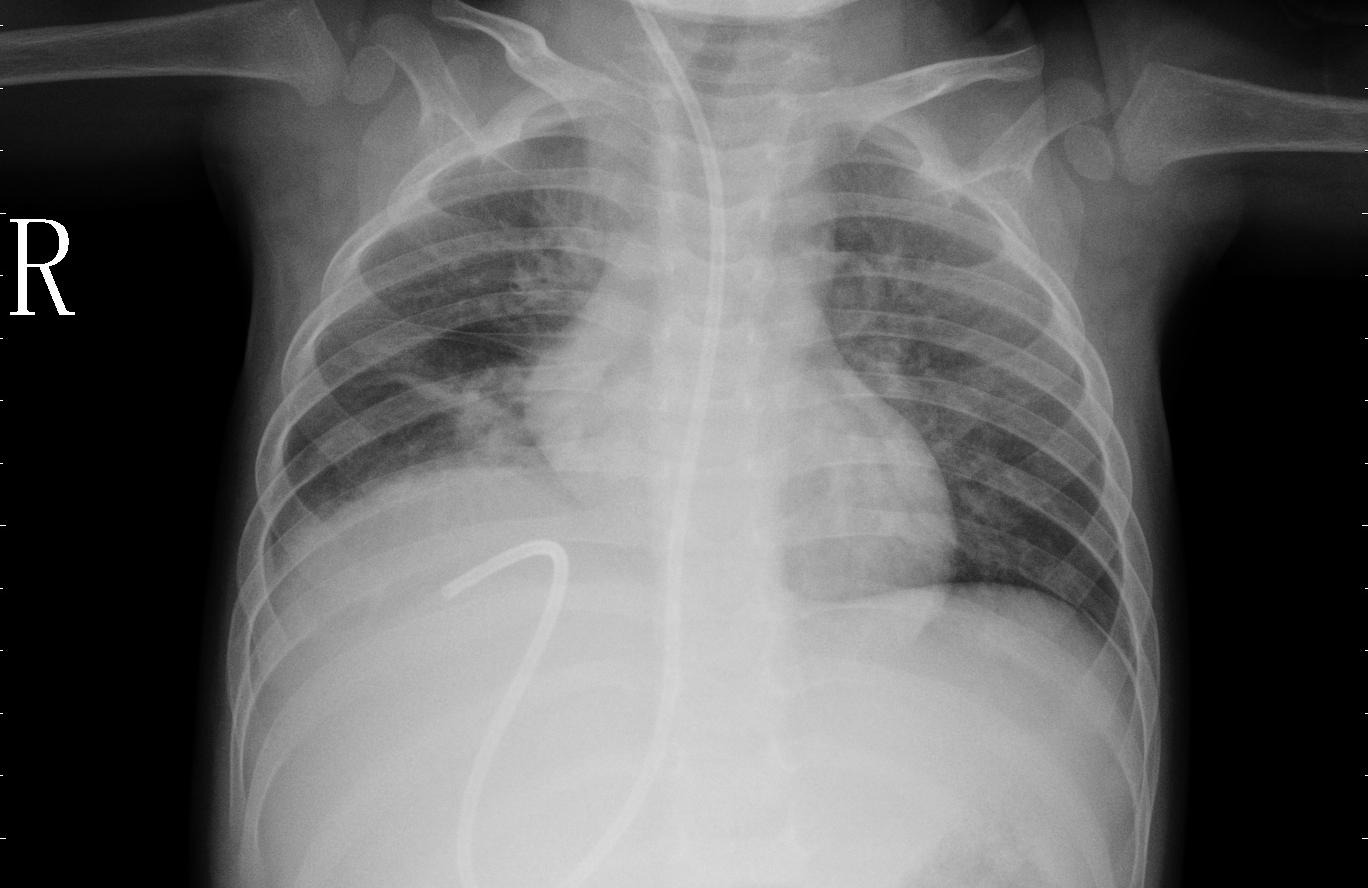}
     \caption{Original Image}
     \label{fig:acc1:high}
    \end{subfigure}
    \begin{subfigure}{0.2\textwidth}
     \includegraphics[width=\linewidth]{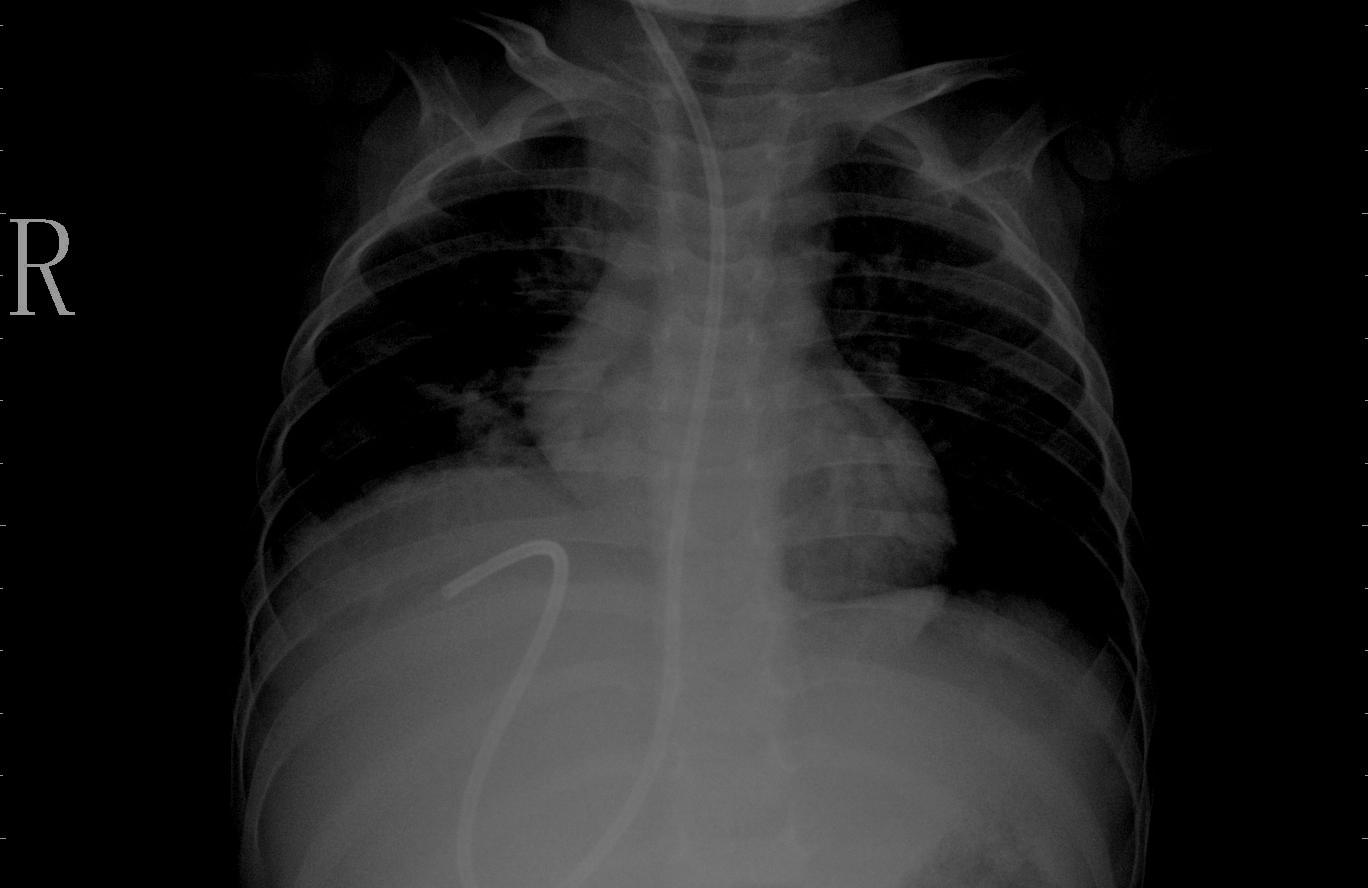}
     \caption{$\beta$ = 1}
     \label{fig:acc1:non}
    \end{subfigure}
    \begin{subfigure}{0.2\textwidth}
     \includegraphics[width=\linewidth]{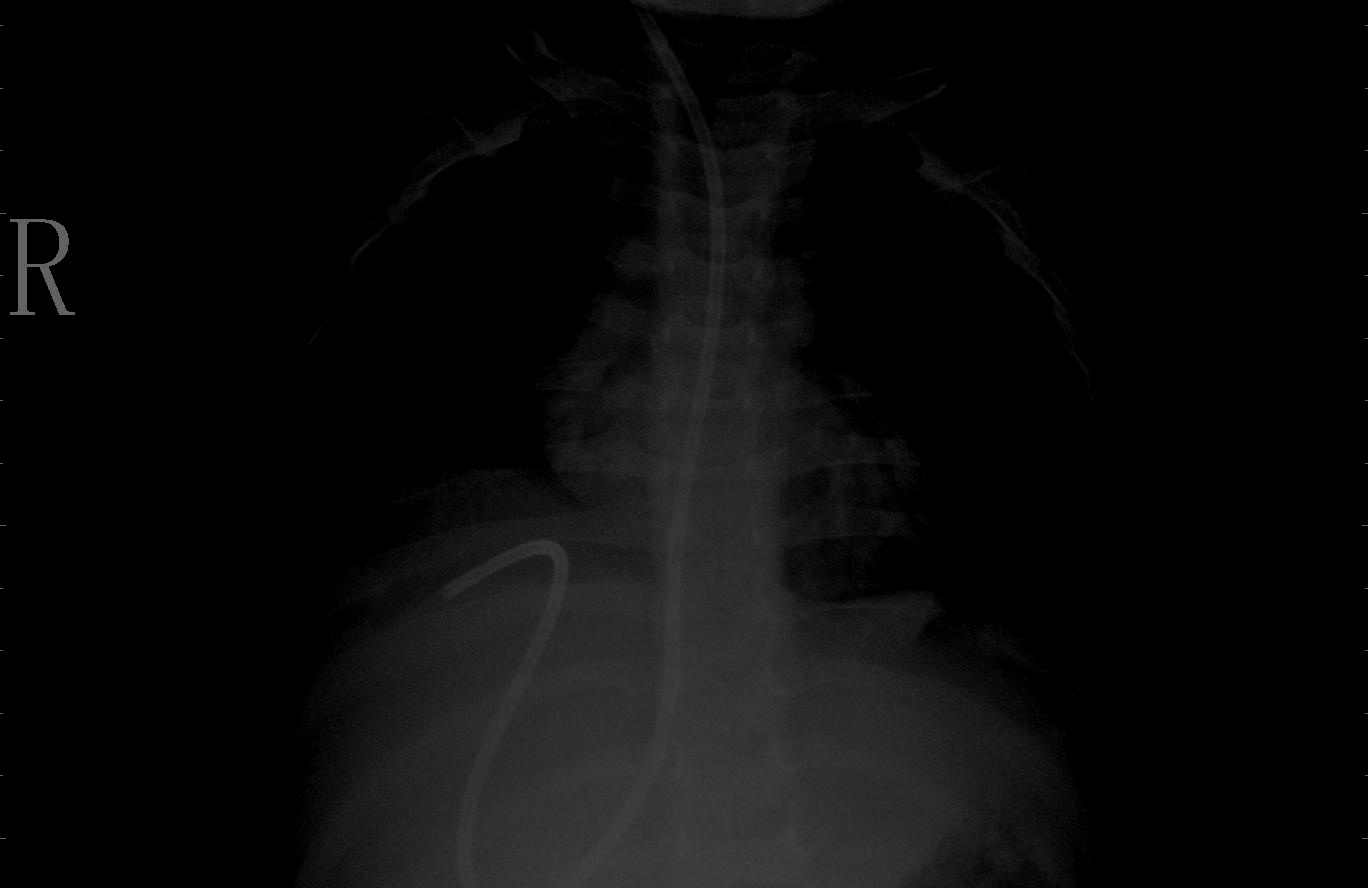}
     \caption{$\beta$ = 2}
     \label{fig:acc1:low}
    \end{subfigure}
    \begin{subfigure}{0.2\textwidth}
     \includegraphics[width=\linewidth]{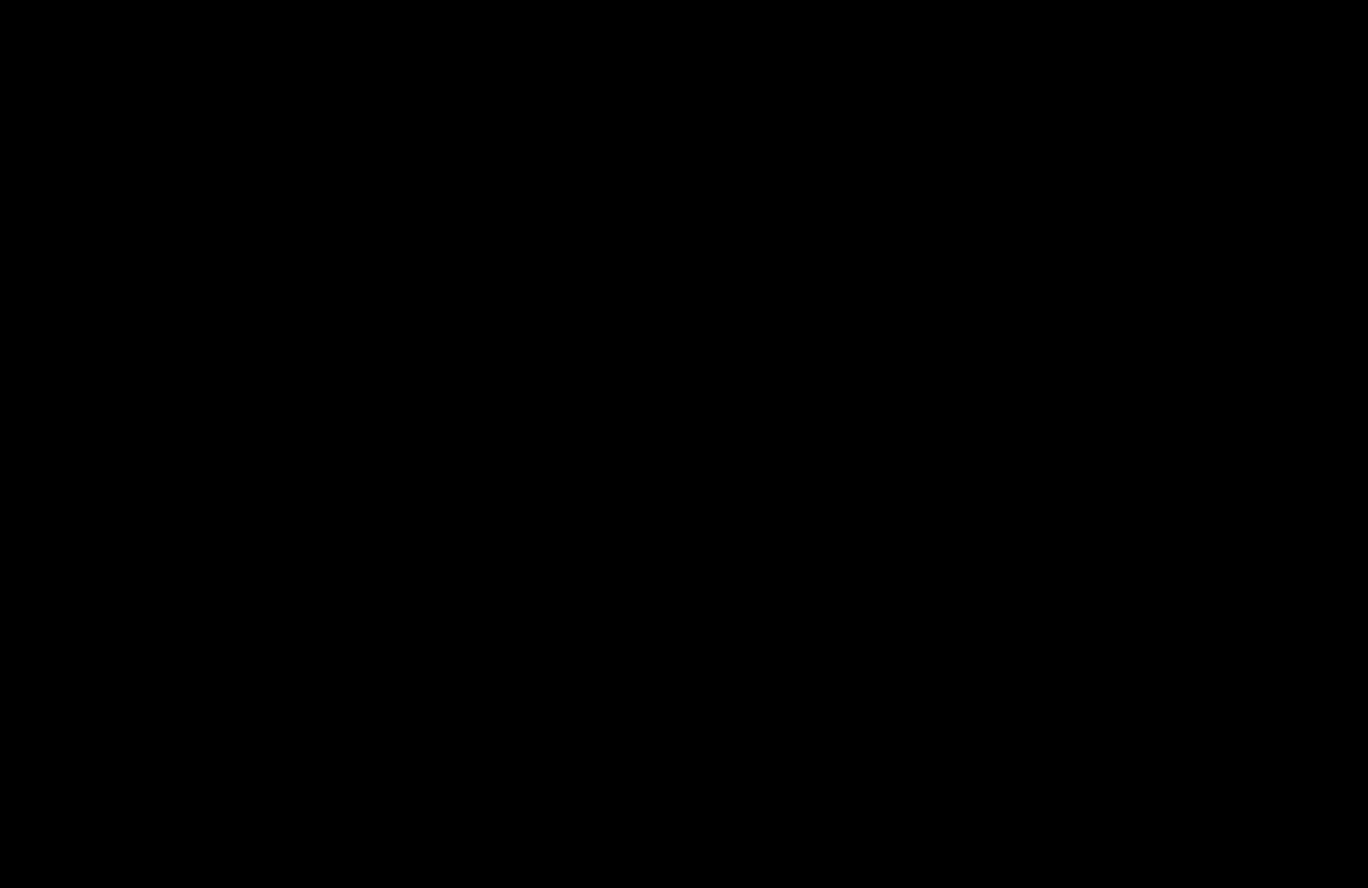}
     \caption{$\beta$ = 4}
     \label{fig:acc1:high}
    \end{subfigure}
    \vspace{-1ex}
    \caption{Comparison of sample image from dataset before and after Local DP based Obfuscation.}
    \label{fig:acc1}
    \vspace{-1ex}
\end{figure*}

%%%%%%%%%%%
\subsection*{Differentially Private Stochastic Gradient Descent (DP-SGD). \cite{abadi2016deep}}
Stochastic Gradient Descent (SGD) is an iterative method for optimizing differentiable objective functions. It updates weights and biases by calculating the gradient of a loss function on small batches of data.
DP-SGD is a modification of the stochastic gradient descent algorithm which provides provable privacy guarantees. It is different from SGD because it bounds the sensitivity of each gradient and is paired with a moments accountant algorithm to amplify and track the privacy loss across weight updates.

In order to ensure SGD is differentially private (i.e. DP-SGD), there are two modifications to be made to the original SGD algorithm. 
First, the sensitivity of each gradient must be bounded. This is done by clipping the gradient in the $l2$ norm. 
Second, one applies random noise to the earlier clipped gradient, multiplies its sum by the learning rate, and then uses it to update the model parameters.

\subsection*{Laplace Distribution. \cite{dwork2014algorithmic}}
The Laplace distribution, also known as the double-exponential distribution, is a symmetric version of the exponential distribution.
The distribution centered at 0 (i.e. $\mu=0$) with scale $\beta$ has the following probability density function:
\begin{center}
$Lap(x|b) = \frac{1}{2\beta} \; \exp(\frac{-|x|}{\beta}) $
\end{center}

The variance of this distribution is $\sigma^2 = 2\beta^2$.

\subsection*{Laplace Mechanism. \cite{dwork2006calibrating, dwork2014algorithmic}}
Laplace Mechanism independently perturbs each coordinate of the output with Laplace noise (from the Laplace distribution having mean zero) scaled to the sensitivity of the function.

Given $\epsilon  \geq 0$ and a target function $f$, the Laplace Mechanism is the randomizing algorithm 
\begin{center}
$A_f(D) = f(D) + x$
\end{center}
where x is random variable drawn from a Laplace distribution , corresponding to perturbation. $\Delta f$ corresponds to the global sensitivity of function $f$, defined as over all dataset pairs that differ in only one element $(D, D')$. 

\begin{figure}[t]
  \centering
  \includegraphics[width=\linewidth, keepaspectratio]{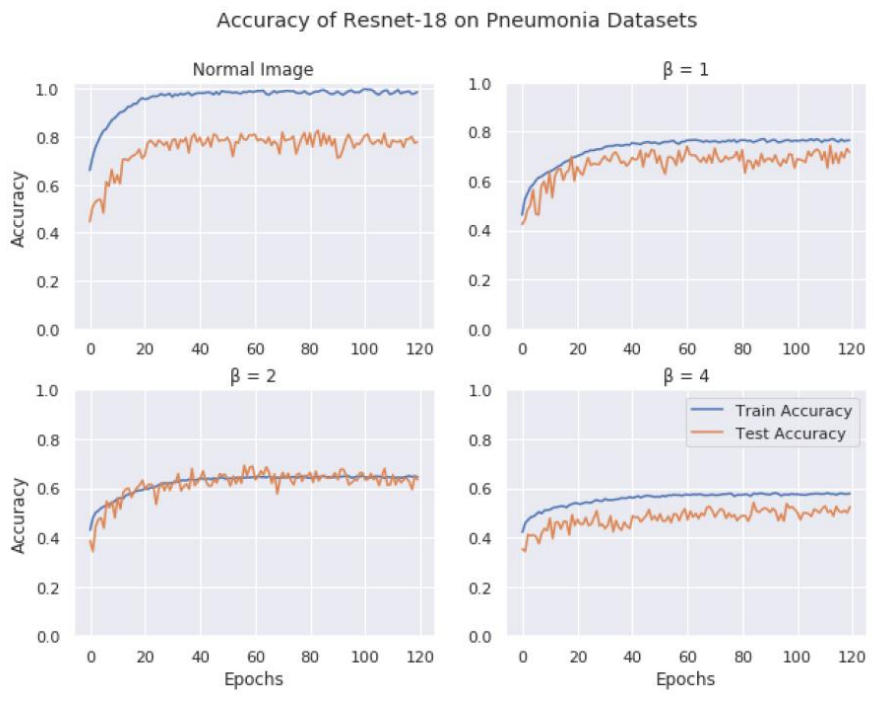}
  \caption{Accuracy vs Epochs for Pneumonia Detection with LDP}
  \label{fig:curve}
  \vspace{-1ex}
\end{figure}

\section*{Experimental Result}
The experiments discussed in this section used an 18-Layer Residual Network (ResNet) previously trained to achieve convergence on the ImageNet task. 
ResNets share many ideas with the popular VGG architecture \cite{simonyan2014very,szegedy2017inception}, with significantly fewer filters and overall decreased complexity. They make use of identity connections between sets of layers as a solution to the common problem of gradient signals vanishing during backpropagation in very deep networks \cite{he2016deep}.
To train the model on these images, some pre-processing steps were undertaken. Input images passed to the deep neural network were scaled to $256\times256$ pixels, and normalized to 1. Therefore, the function $f$ is the identity function and the sensitivity $\Delta f$ is 1. 
The experiments are all carried out using Python 3.8.2 and PyTorch 1.4.0.

For the process of Local-DP, 3 other versions of datasets were generated by the addition of different perturbations to the images. These alternate, differentially-private datasets were generated by drawing random samples from the Laplacian Mechanism mentioned in \cite{dwork2006calibrating} with $\mu =0$ and varying levels of scale i.e. $\beta $. These perturbations were added directly to the input image to create a noisy representation for subsequent training. These datasets were used in different experiments to train the Resnet-18 model to convergence.
We trained separate instances of the pre-trained Resnet-18 model on these 3 image datasets, over 50 epochs. The best models from these runs were saved and analyzed. The tradeoff in accuracy with varying scales of perturbations ($\beta =1; \beta=2; \beta =4$) were examined. 

In the case of DP-SGD, we make two main modifications to the original SGD algorithm. We clip the gradient in the $l2$ norm, add random noise to it and then multiply it by the learning rate before updating the model parameters. In a similar manner to the LDP experiments, the best models from these runs were saved and analyzed. The tradeoff in accuracy with varying scales of noise ($\beta =1; \beta=2; \beta =4$) were examined.

%%%%%%%
\begin{figure*}
  \centering
  \includegraphics[width=\linewidth, keepaspectratio]{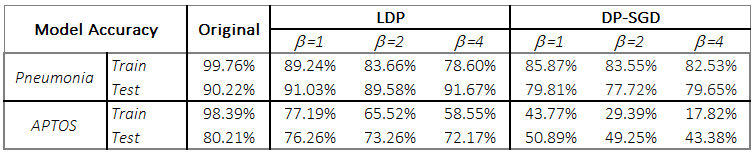}
  \caption{Model Accuracy (\%) | DP mechanism vs Dataset}
  \label{fig:model}
  \vspace{-1ex}
\end{figure*}
%%%%%%%%%%%

\newpage

\subsection*{Pneumonia Detection}

\textbf{Setup}: 

The experimental setup consisted of training the selected model on an image classification task on the Chest X-Ray dataset \cite{kermany2018identifying}. The LDP experiments followed the process outlined earlier, and resulted in 3 noisy datasets. Examples of perturbed images can be seen in Figure \ref{fig:acc1}. The DP-SGD experiments involved the modifications outlined earlier, including gradient clipping and multiplication by the learning rates prior to model parameter updates. The ResNet-18 model has around 11M parameters and was pre-trained on ImageNet and trained using the SGD optimizer, 0.01 learning rate, and batch size of 128. The network was run over 50 Epochs. 

\textbf{Results}: (Figure \ref{fig:model}) Best model accuracies on both the train and test set from the 50 epoch runs were saved. On the original Chest X-Ray dataset, the model achieved 99.76\% accuracy on training data and 90.22\% on the held out test set. We observe that training accuracy drops significantly, in the case of both LDP and DP-SGD, as we increase $\beta$ from 1 to 4. On the other hand, we see that the test accuracy doesn't reflect this trend. Results for a longer run of 120 epochs are also included in Figure \ref{fig:curve} as learning curves.

% \textbf{Results}: Best model accuracies on both the train and test set from the 50 epoch runs were saved. On the original Chest X-Ray dataset, the model achieved 99.76\% accuracy on training data and 90.22\% on the held out test set. With images modified through LDP by perturbations of scale $\beta = 1$, the network achieved 89.24\% and 91.04\% accuracy on the train and test set, respectively. For LDP with $\beta = 2$, the network achieved 83.66\% and 89.58\% accuracy on the train and test set, and for $\beta = 4$ it achieved 78.60\% and 91.67\% accuracy. Results for a longer run of 120 epochs are also included in Figure 3 as learning curves. For DP-SGD, at $\beta = 1$ the network achieved 85.87\% and 79.8\% accuracy on the train and test set, respectively.
% For $\beta = 2$, the network achieved 83.55\% and 77.72\% accuracy on the train and test set, and for $\beta = 4$ it achieved 82.53\% and 79.65\% accuracy. 

\subsection*{Blindness Detection}

\textbf{Setup}: This experimental setup consisted of training the model on the APTOS dataset described earlier. The simulations were structured in the same way for both LDP and DP-SGD, using the same Resnet-18 model previously trained on Imagenet. Preprocessing, postprocessing, and evaluation steps were identical to the Pneumonia Detection experiments.

\textbf{Results}: (Figure \ref{fig:model}) As before, the best model accuracies on both the train and test set from the 50 epoch runs were saved. On the original APTOS dataset, the model achieved 98.39\% accuracy on training data and 80.21\% on the held out test set. Similar to the behaviour in the Pneumonia detection task, we observe that training accuracy drops significantly in the case of both LDP and DP-SGD as we increase $\beta$ from 1 to 4, while the test accuracy doesn't reflect this trend.

\section*{Conclusion}
In this paper, we provided an empirical evaluation of a fine tuned Resnet model trained on medically relevant classification tasks. We observe that the DP noise mechanisms lead to varied results with different perturbation levels and highlights the inherent trade-offs in these decisions. As expected, we notice the model accuracy declines as we increase the level of privacy guarantees ($\beta$). 

We noticed earlier in the experiments that while the training accuracy dropped significantly as we increased $\beta$, the test accuracy doesn't change much. We also observe that in some cases the training accuracy is lower than the test accuracy (Figure \ref{fig:model}). One possible explanation for this is that Local-DP adds noise to the training data, making the latent features harder to learn. Later when we run the model on the test data it performs better because the latent features are now relatively easy to capture since the model has already learned representations in a noisy scenario. 

We observe that while Local-DP maintains the theoretical guarantee of Differential Privacy, it does not always provide the visual privacy we expect. This could be a problem during the inference stage where passing an image with all sensitive features intact could cause security breaches.
This provides an interesting future direction to experiment with more Local-DP techniques, particularly those which take into account sensitive features. Another interesting research direction includes benchmarking differential privacy mechanisms across various data modalities including audio files, video, text, tabular data, and various forms of cyber data.

\newpage

\bibliography{example_paper}
\bibliographystyle{icml2020}
\end{document}